**Title**: Semi-supervised segmentation of tooth from 3D Scanned Dental Arches

Ammar Alsheghri [1], Farnoosh Ghadiri [2], Ying Zhang [1], Olivier Lessard [1], Julia Keren [3], Farida Cheriet [1], Francois Guibault [1]

1- Department of Computer and Software Engineering, École Polytechnique de Montréal, Montréal, QC, Canada

2- Centre d'intelligence artificielle appliquée (JACOBB), Montréal, QC, Canada

3- Intellident Dentaire Inc., Montréal, QC, Canada

**Abstract**

Teeth segmentation is an important topic in dental restorations that is essential for crown generation, diagnosis, and treatment planning. In the dental field, the variability of input data is high and there are no publicly available 3D dental arch datasets. Although there has been improvement in the field provided by recent deep learning architectures on 3D data, there still exists some problems such as properly identifying missing teeth in an arch. We propose to use spectral clustering as a self-supervisory signal to joint-train neural networks for segmentation of 3D arches. Our approach is motivated by the observation that K-means clustering provides cues to capture margin lines related to human perception. The main idea is to automatically generate training data by decomposing unlabeled 3D arches into segments relying solely on geometric information. The network is then trained using a joint loss that combines a supervised loss of annotated input and a self-supervised loss of non-labeled input. Our collected data has a variety of arches including arches with missing teeth. Our experimental results show improvement over the fully supervised state-of-the-art MeshSegNet when using semi-supervised learning. Finally, we contribute code and a dataset.

**Summary:** Dental offices tackle thousands of dental reconstructions every year. Complexity and abnormalities in dentition make segmentation of an optical scan a challenging manual task that takes 45 minutes on average. The present work improves the generalization of currently available deep learning segmentation model (MeshSegNet) on 3D dental arches by introducing a new loss function to leverage unlabeled available data. The segmentation network is trained using a joint loss that combines a supervised loss of annotated input and a self-supervised loss of non-labeled input. Our experimental results show improvement over the fully supervised state-of-the-art MeshSegNet when using semi-supervised learning.

1. **Introduction**

Semantic segmentation in dentistry implies separating teeth from each other and from the surrounding gingiva and performing classification of tooth type (Figure 1). Teeth segmentation is important for crown generation, diagnosis, and treatment planning [3]. It can be achieved manually using commercial dental software, but it is time-consuming [3]. Hence, achieving this objective accurately and automatically is still an open area of research for restorations, where the variability of scanned arches (i.e., input data) is high.

There are many studies on segmentation of non-Euclidean data [1, 2, 5, 6]. Among them, the MeshSegNet neural network which processes dental arches as point clouds and provides state-of-the-art results for teeth segmentation [2]. However, the MeshSegNet was trained on ideal arches that contain 14 teeth, which is not always the case in real life scenarios. Particularly, for our data, MeshSegNet lacks generalization for cases with missing teeth, irregular teeth, or preparation.

The goal of this study is to increase performance of the tooth segmentation out of a fixed supervised model by introducing a new self-supervised loss function. The dataset and code are publicly available at https://github.com/Alsheghri/Teeth-Segmentation.

2. **Methods**

*2.1 Materials:* Our dataset contained 24 arches in total and was split as follows:
- Supervised training data: 4 labeled arches containing missing teeth at different locations.
- Self-supervised training data: 14 unlabeled arches; 9 of them had missing teeth at different locations.
- Testing data: 6 arches; 3 of them had missing teeth at different locations.

The arches used in the self-supervised training had cases with irregular teeth and cases with partial arches.

*2.2 Data preprocessing:* The supervised training and test data were manually segmented by a technician using Exocad to produce labeled data. Exocad is a commercial dental software used by several dental laboratories worldwide to design dental restorations. Regular augmentation methods such as rotation, translation, and scaling in reasonable ranges were applied on the decimated meshes before passing them for training. The coordinated of all arches were normalized by subtracting the mean of the points and dividing by the standard deviation.

*2.3 Network Architecture:* We used the MeshSegNet architecture [2] as a baseline for supervised training which is based on pointNet++ [5] and includes additional features to improve teeth segmentation. We explored using a self-supervised training loss of unlabeled data to train the network and improve the segmentation performance. The adopted network architecture is presented in Figure 1.

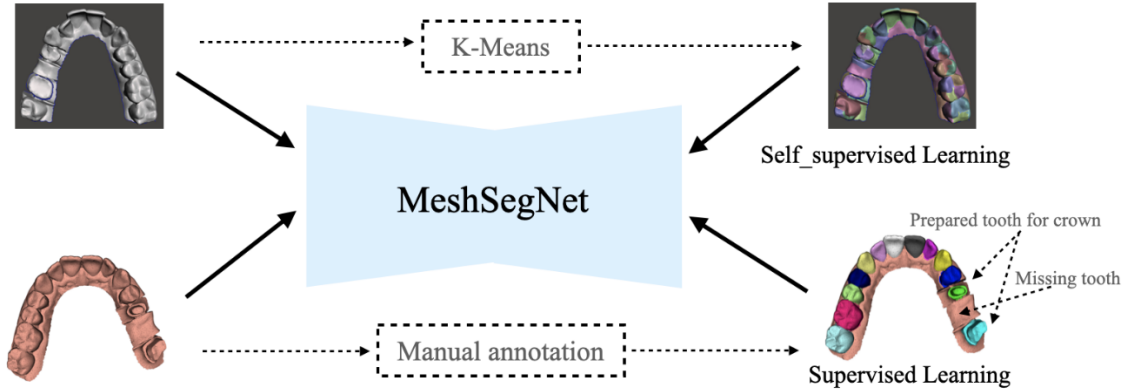

Figure 1: Overview of our method. Top: K-means clustering is applied on a subset of our dataset which is unlabeled, yielding a self-supervised training signal for the neural network; Bottom: fully supervised setting, where the segmentation of mesh is provided manually using Exocad, which is then used as supervision for the neural network.

K-means clustered arches were used by the self-supervised learning method [4]. Given a set of 3D mesh faces, an affinity matrix which encodes the likelihood of each pair of faces belonging to the same group was first constructed. The eigenvectors of the affinity matrix were used to obtain data representations and enhance clustering of data points in the embedding space [4]. A parametric study was performed to fine-tune the model parameters and enhance the clustering of arches. In particular, the number of clusters was set to 60, the parameter $\delta$ which controls the relative importance of geodesic distance and angle distance was set to 0.03, the parameter $\eta$ which controls the weight of concavity was set to 0.15, and k-mean++ was used for K-means initialization. Each arch was clustered into 60 clusters to ensure that no cluster could contain parts from a tooth and gingiva at the same time. We generated training data by decomposing unlabeled 3D arches into smaller components as shown in Figure 2. We noticed that K-means clustering decomposes the arches into segments that respect the presence of margin lines between teeth and gingiva. The model was then trained using a joint contrastive loss as inspired by [1].

*2.4 Semi-supervised learning with K-means clustering:* We leveraged the use of self-supervised loss to benefit from unlabeled unseen dental arches. This approach could be used to leverage information from unlabeled non-ideal cases such as partial arches and arches with irregular teeth as shown in Figure 2.

Consider samples $X = \{x_i\}_{i \in [n]}$ divided into $X^L$ and $X^U$ of sizes $l$ and $u$, respectively. $X^L$ consists of point clouds that are provided with human-annotated labels $y^L$, while we do not know the labels of the samples $X^U$. By clustering the samples in $X^U$, we can obtain a set of components for each shape. The pairwise contrastive loss $L^{self-supervised}$ is defined over $x_i \in X^U$ as a self-supervised objective. On the other hand, the generalized dice loss $L^{supervised}$ is used for sample $X^L$ because the ground-truth labels are provided. A contrastive loss can be defined such that points belonging to the same component have a high similarity and points from different components have low similarity as inspired by [1]. Therefore, the network was trained on both $X^L$ and $X^U$ via a joint loss $L$ that combines the supervised loss and the self-supervised loss with a hyper-parameter $\lambda = 10$ to control the relative strength between the losses such that:

$$L = L^{supervised} + \lambda . L^{self-supervised} \quad Eq.(1)$$

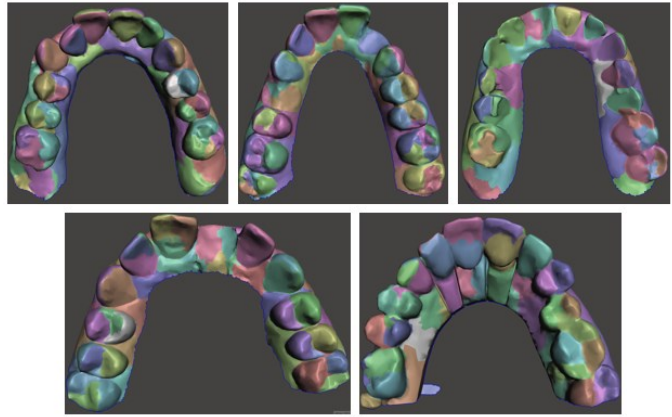

Figure 2: K-means clustering with 60 clusters for some arches used in self-supervised training showing that margin lines between teeth and gingiva are visible between different clusters.

## 3. Results

### *3.1 The effect of using self-supervised loss on unlabeled data*

To highlight the effect of semi-supervised learning, we performed an experiment to compare two different models; one with semi-supervised training that included a self-supervised loss and the other with purely supervised training as in MeshSegNet [2]. The purely supervised model was only trained on the supervised data of 4 arches. The semi-supervised model was trained on the same supervised data and on the self-supervised data of 14 unlabeled arches (see Figure 3).

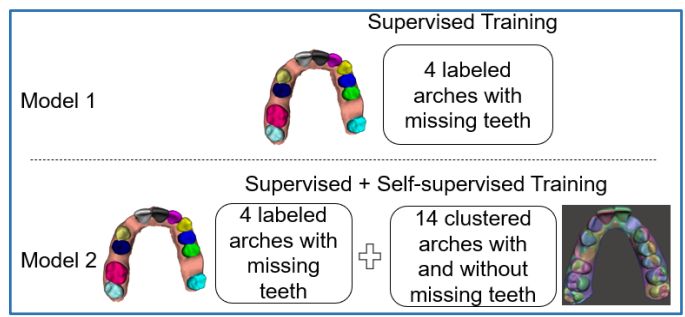

Figure 3: Training data for supervised and semi-supervised models.

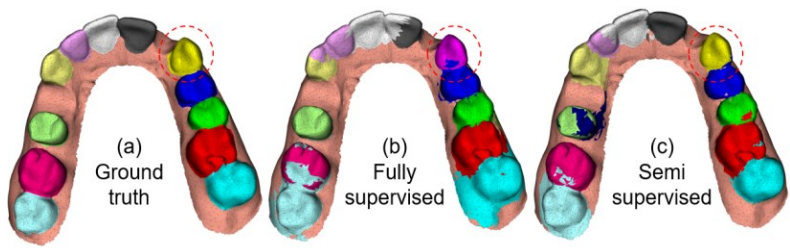

Figure 4: Results showing the effect of adding a self-supervised loss while training with the same amount of labeled data. (a) ground-truth; (b) segmentation of test case using supervised model; (c) segmentation of test case with semi-supervised model. The purple label in subfigure (b) is incorrect as compared with ground truth.

The results show that combining self-supervised and supervised learning improves the results compared with purely supervised learning (Figure 4). The semi-supervised training increased the Dice Similarity Coefficient by 13 % compared with supervised training alone for the same amount of supervised labeled data as shown in Table 1. The table shows that the individual DSC scores for all test cases are higher in the case of training with a self-supervised loss compared with training without self-supervised loss. In general, the scores for arches without missing teeth are greater than those for arches with missing teeth, which caused the large variances observed in both groups.

Table 1: Results for fully supervised and semi-supervised models tested on 6 unseen dental arches; 3 of them had missing teeth at different locations. Dice similarity coefficients (DSC) are reported per arch, as well as the average over all arches.

| Arch # | Arch has missing teeth? | DSC for training without self-supervised loss | DSC for training with self-supervised loss |
|---|---|---|---|
| 1 | no | 0.71 | 0.89 |
| 2 | no | 0.62 | 0.86 |
| 3 | no | 0.76 | 0.85 |
| 4 | yes | 0.60 | 0.68 |
| 5 | yes | 0.44 | 0.56 |
| 6 | yes | 0.70 | 0.76 |
| Average | | $0.64 \pm 0.12$ | $0.77 \pm 0.13$ |

**4. New Work to be presented**

We proposed a new loss by adding the loss of self-supervised training using unlabeled data to the loss of supervised training using labeled data, which improved the performance of the 3D tooth segmentation model. Code and dataset will be available publicly.

**5. Conclusions**

It is concluded that combining representations obtained from self-supervised learning with supervised learning improves the generalization of the trained deep learning model for 3D tooth segmentation in the case of few available labeled data.

**6. Acknowledgments**

This work was funded by KerenOr, Intellident Dentaire Inc., iMD Research, the Natural Science and Engineering Research Council of Canada, Institut de valorisation des données (IVADO), and MEDTEQ. We thank Compute Canada for providing the computational resources used in this work. The authors acknowledge the help and support from JACOBB and Object Research Systems Inc. This work has not been submitted for publication or presentation elsewhere.